\documentclass[10pt,twocolumn,letterpaper]{article}

\usepackage{iccv}
\usepackage{times}
\usepackage{epsfig}
\usepackage{graphicx}
\usepackage{amsmath}
\usepackage{amssymb}
\usepackage{multirow}

\usepackage[breaklinks=true,bookmarks=false]{hyperref}

\iccvfinalcopy 

\def\iccvPaperID{****} 

\begin{document}

\title{Fast and Full-Resolution Light Field Deblurring using a Deep Neural Network}

\author{Jonathan Samuel Lumentut$^1$\\
{\tt\small jlumentut@gmail.com}
\and
Tae Hyun Kim$^2$\\
{\tt\small lliger9@gmail.com}
\and
Ravi Ramamoorthi$^3$\\
{\tt\small ravir@cs.ucsd.edu}
\and
In Kyu Park$^1$\\
{\tt\small pik@inha.ac.kr}
\and
$^1$Inha University\\
\and
$^2$Hanyang University\\
\and
$^3$University of California, San Diego\\
}

\maketitle

\begin{abstract}
Restoring a sharp light field image from its blurry input has become essential due to the increasing popularity of parallax-based image processing. State-of-the-art blind light field deblurring methods suffer from several issues such as slow processing, reduced spatial size, and a limited motion blur model. In this work, we address these challenging problems by generating a complex blurry light field dataset and proposing a learning-based deblurring approach. In particular, we model the full 6-degree of freedom (6-DOF) light field camera motion, which is used to create the blurry dataset using a combination of real light fields captured with a Lytro Illum camera, and synthetic light field renderings of 3D scenes. Furthermore, we propose a light field deblurring network that is built with the capability of large receptive fields. We also introduce a simple strategy of angular sampling to train on the large-scale blurry light field effectively. We evaluate our method through both quantitative and qualitative measurements and demonstrate superior performance compared to the state-of-the-art method with a massive speedup in execution time. Our method is about 16K times faster than Srinivasan et. al.~\cite{Srinivasan_CVPR2017} and can deblur a full-resolution light field in less than 2 seconds.
\end{abstract}

\vspace{-5mm}
\section{Introduction}
In the last decade, 2-dimensional (2D) image deblurring problem has been a popular topic in computer vision with the specific trends such as fast and robust processing~\cite{Krishnan_CVPR2011, Pan_CVPR2016, Whyte_CVPR2010, Xu_ECCV2010}.
Unlike the conventional 2D image representation, a 4D light field (LF) image contains both spatial and angular information to represent a pixel value, and the pixels collected from a specific angular direction form a specific 2D sub-aperture image.
Thus, it require different approaches compared to conventional 2D image processing.
Recently, consumer-oriented LF cameras have been developed by Lytro~\cite{Lytro1} and Raytrix~\cite{Raytrix1}, and multi-camera setups are increasingly becoming popular in modern smartphones.
By utilizing LF cameras, there have been significant advances in many computer vision tasks such as depth-estimation~\cite{Wang_TPAMI2016, Williem_TPAMI2018}, refocusing effect~\cite{Ng_TOG2015}, and view synthesis~\cite{Kalantari_TOG2016, Srinivasan_ICCV2017}.

\begin{figure}
\begin{center}
		\includegraphics[width=0.45\textwidth]{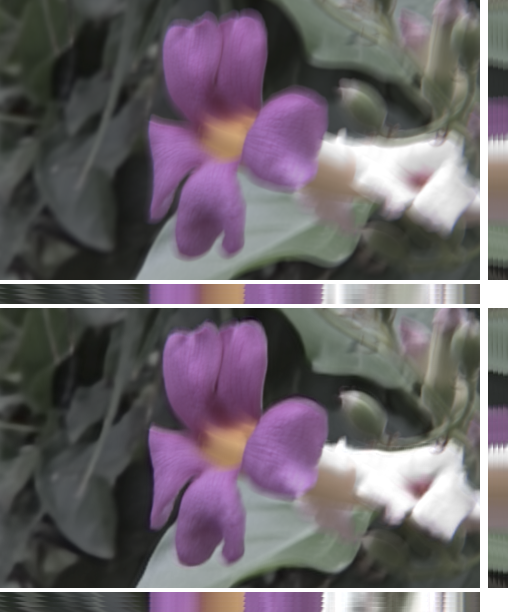}
\end{center}
\vspace{-0.2cm}
\caption{\textbf{First Row}: Central sub-aperture image of blurry light field. \textbf{Second Row}: Central sub-aperture image of deblurred light field from our result. Our proposed network performs full spatio-angular resolution light field deblurring within 2 seconds and renders a high-quality 4D light field image. Note the 2D slice of the EPIs shown which indicates full light field deblurring.}
\label{fig:Show_up_fig}
\end{figure}
Surprisingly, post-capture deblurring is rarely done on LF images.
As opposed to simple image deconvolution, the blur in LF images is often depth-dependent.
The closest previous work is that of Srinivasan~\etal~\cite{Srinivasan_CVPR2017}.
In the general case, they apply a full optimization approach to 4D spatio-angular data to estimate motion blur.
However, this approach is limited with several issues such as high memory requirement, slow processing, and limited LF resolution.
Their deblurring model is also limited to 3-DOF translation parameters, and does not represent real world motion blur.
These problems were solved partially by recent LF deblurring works~\cite{Dongwoo_ECCV2018, Mohan_CVPR2018} but are still inapplicable on any LF camera as the post-capture processing, due to their slow execution time ($\sim$30 minutes).

In this paper, we introduce a fast LF deburring network which is trained using a blurry LF dataset.
To generate the large blurry LF dataset for training the network, we assume general 6-DOF (3-DOF translation + 3-DOF rotation) camera motion model to simulate the realistic motion blur.
Our method is designed to address the limitation of previous works that assume 3-DOF translational~\cite{Srinivasan_CVPR2017} and 3-DOF motion density function (MDF)~\cite{Mohan_CVPR2018}.
Based on the general motion blur model, we build a synthetic LF blur dataset and train the proposed network by using the acquired large dataset.
In our LF deblurring network, we combine convolution and deconvolution layers with recurrent procedure to produce robust LF deblurring results.
The proposed network is equipped with recurrent network and being trained multiple times to increase its receptive field capability without a very deep architecture, and we can carry out a very fast, consistent, and full-resolution LF deblurring.
A direct observation of our result is shown in Figure~\ref{fig:Show_up_fig}.
Finally, we conduct a quantitative and qualitative comparison using our test-set and real LF images to evaluate the performance of the network.
In summary, our contributions are:
\begin{itemize}
\vspace{-0.2cm}
\item We present a 6-DOF camera motion blur model for LF image motion deblurring. $\triangleright$~{Sec.~\ref{full_6DOF_model_sec}}
\vspace{-0.2cm}
\item We provide a novel 6-DOF LF blur dataset along with the ground truth data for training and benchmarking the LF deblurring algorithms. $\triangleright$~{Sec.~\ref{Blur_dataset_sec}}
\vspace{-0.2cm}
\item We propose, to the best of our knowledge, the first neural network based LF deblurring that shows its superiority in small computational time as well as its capability of processing full-resolution LF. $\triangleright$~{Sec.~\ref{LFRDBN_sec}}.
\end{itemize}
\section{Related Work}
\noindent{\bf Blind Image Deblurring}
Blind image deblurring is a method used to recover a sharp image from a blurry input by estimating the point spread function (PSF) or blur kernel that records the camera shake.
Previous work by Whyte~\etal~\cite{Whyte_CVPR2010} introduced a non-uniform deblurring method by projecting blurred images under homography from a 3-DOF rotation.
Similarly, Tai~\etal~\cite{Tai_TPAMI2011} generalized the deblurring problem by modeling the blur output as a sequence of planar projective transformations of a sharp image.
Studies on image priors are also introduced to refine the deblurring result.
Levin~\etal~\cite{Levin_TOG2007} introduced a sparse derivative prior that concentrates on the derivatives of low intensity pixels.
This approach is aimed at solving defocus blur, but also applicable for motion-blurred images.
Xu and Jia~\cite{Xu_ECCV2010} estimated a sparse blur kernel by preserving edge information in a similar way as a hysteresis method but with different threshold settings.
Krishnan~\etal~\cite{Krishnan_CVPR2011} introduced a regularization ratio to estimate the best blur kernel.
Recent prior such as intensity distribution by Pan~\etal~\cite{Pan_CVPR2014} is also among the best performer in deblurring large blurs as discussed in a recent comparative study~\cite{Lai_CVPR2016}.
Pan~\etal~\cite{Pan_CVPR2016} improved their work and they showed robust performance in text deblurring.
Recent blind image deblurring algorithms utilized deep neural network techniques.
These recent works apply image deblurring directly without blur kernel estimation~\cite{Kim_ICCV2017, Kupyn_CVPR2018, Nah_CVPR2017, Su_CVPR2017}.
The 2D image deblurrings may not, however, be applicable on 4D LF image due to the different camera model.
\\
\\
\noindent{\bf Light Field Deblurring}
Recent works on LF deblurring consist of blind and non-blind approaches.
Dansereau~\etal~\cite{Dansereau_CVPRW2017} utilized the traditional Ricardson-Lucy deblurring on LF images.
Even though their experiment showed superior performance, this approach still requires PSF information, unlike blind deblurring.
State-of-the-art work by Srinivasan~\etal~\cite{Srinivasan_CVPR2017} estimates blurs on LF under 2-DOF in-plane and 1-DOF out-of-plane translation (3-DOF).
Their work excludes the need for rotation motion, which is common to real motion-blurred image~\cite{Whyte_CVPR2010}.
This problem is solved by Mahesh Mohan and Rajagopalan~\cite{Mohan_CVPR2018} who implemented 2-DOF in-plane translation and 1-DOF $z$-axis rotation model (3-DOF) following the model of motion density function (MDF)~\cite{Gupta_ECCV2010}.
Although their model was able to produce a better result than the state-of-the-art~\cite{Srinivasan_CVPR2017}, the MDF model did not include out-of-plane translation ($z$-axis translation).
The newer approach of LF deblurring is introduced by Lee~\etal~\cite{Dongwoo_ECCV2018}.
This approach performs pose estimation on each sub-aperture image to warp the estimated blur kernel for deblurring the entire 4D LF.
This method, however, also shows some limitations such as slow run-time and non-complex motion blur constraint.

Unlike the previous approaches, we are motivated to solve LF deblurring using a deep neural network. Earlier works have achieved state of the art performance for deblurring 2D images and 3D videos~\cite{Kim_ICCV2017, Nah_CVPR2017, Su_CVPR2017}.
In this work, we generate the full 6-DOF model of LF camera motion and implement a tool to synthesize complex blurry LF dataset from the given sharp input.
The blur model is designed within 6-DOF motion as opposed to the 3-DOF model from previous approaches~\cite{Mohan_CVPR2018, Srinivasan_CVPR2017}.
Finally, we generate a deep neural network for fast and robust LF deblurring.
\section{Proposed Method}
In this section, we discuss the complete framework of the LF blur model and the LF deblurring network.
We start the framework by elaborating the 6-DOF motion model to produce a blurry LF dataset.
We then proceed to the deblurring network that seeks to alleviate both the synthetic and real motion-blurred LF.

\subsection{6-DOF Light Field Blur Model}
\label{full_6DOF_model_sec}
In this work, given sharp sub-aperture images as input, blurry LF sub-aperture images are generated by simulating 6-DOF LF camera motion.
Instead of generating blur kernels to convolve sharp sub-aperture images one by one, we opt to project the sharp sub-aperture images according to the LF camera motion and integrate them over time.
This technique is similar to a previous work~\cite{Tai_TPAMI2011} that warps sharp 2D image into several warped images using different homographies.
The homographies are built based on hand-held conventional camera motion, and the final blur output is produced by averaging those warped images.
Similarly, we model the 6-DOF LF camera motion to synthesize the blurry sub-aperture images.
In the following discussion, we develop the 3-DOF translation model described by an earlier work~\cite{Srinivasan_CVPR2017}, followed by the 3-DOF rotation model.
These can be combined to represent full 6 DOF LF camera motion.
This model is based on the geometry of LF camera motion.
In practice, we apply it in both real LF images and synthetic 3D scenes to generate the blur dataset.

\subsubsection{3-DOF Translational Motion}
The 3-DOF LF translational blur model is approximated by 2D in-plane motion along the $x-$ and $y-$ axis and 1D out-of-plane motion along the $z-$axis.
The LF is considered as a collection of pinhole camera models that are positioned at angular coordinates $(u, v)$.
The blurry LF is produced using a model that integrates all warped and sheared copies of the sharp LF ($S$).
As a result, a blurry pixel ($B$) on a LF can be
\begin{equation}
\label{eq:eq_3DOF_blur}
		B(x,u) = \int_{t} S(x, u + p_x(t) - x p_z(t)) \,dt,
\end{equation}
where ($x, u$) represents the spatial and angular coordinate while $( p_x(t), p_z(t) )$ represents the in-plane and out-of-plane camera translation in an exposure time $t$.
We reduce the 4D LF parameters from ($x, y, u, v$) to ($x, u$) for simplicity, but the extension to full 4D LF is straightforward.

\subsubsection{3-DOF Rotational Motion}
To generate a full 6-DOF blur model, Eq.~(\ref{eq:eq_3DOF_blur}) needs the additional information of 3-DOF rotation parameters.
The rotation on LF camera consists of 2 parts: out-of-plane rotation ($x$-axis (pitch) and $y$-axis (yaw) rotation), and in-plane rotation ($z$-axis (roll) rotation).
We separately discuss the out-of-plane and in-plane rotations model to construct the final formulation of full 6-DOF motion.
\begin{figure}
\begin{center}
		\includegraphics[width=0.45\textwidth]{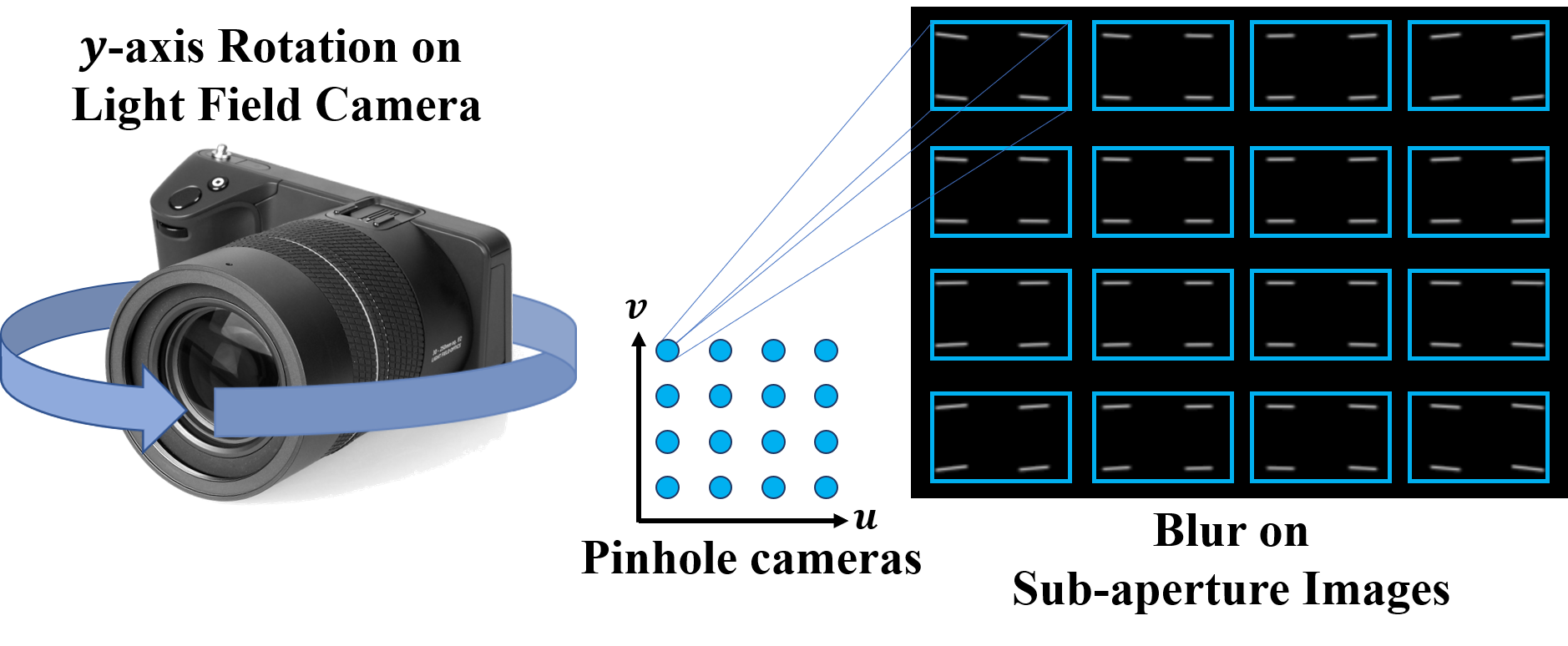}
\end{center}
\vspace{-0.2cm}
\caption{Illustration of the blurred result on sub-aperture images when the LF camera is rotated along the $y$-axis.}
\label{fig:LF_Outplane_Rotation}
\end{figure}
\\
\\
\noindent{\bf Out-of-plane Rotation}
Similar to the 3-DOF translation, we model sub-aperture images as a collection of 2D images captured by multiple pinhole cameras as shown in Figure~\ref{fig:LF_Outplane_Rotation}.
This model allows any fronto-parallel sub-aperture image $S(u)$ to be warped by a $3 \times 3$ homography $\textbf{H}_{u, t}$ that varies along angular position $u$ and exposure time $t$ to produce blurry sub-aperture image $B(u)$.
This term can be written as
\begin{equation}
\label{eq:homog_rot_y_blur}
		B(u) = \int_{t} \textbf{H}_{u,t}S(u)\,dt,
\end{equation}
where its homography can be constructed by the following
\begin{equation}
\label{eq:homog_eq}
\textbf{H}_{u,t} = \textbf{K}_u \textbf{R}_t \textbf{K}_u^{-1}.
\end{equation}
In above expression, $\textbf{R}_t$ is the $3\times3$ rotation matrix that varies along $t$ and $\textbf{K}_u$ is the intrinsic matrix that varies along the pinholes' position, $u$ (angular domain).
The $\textbf{R}_t$ is defined using Rodrigues’ rotation matrix as
\begin{equation}
\label{eq:rot_exponential}
	\textbf{R}_{t} = e^{\textbf[\Omega_{t}]},
\end{equation}
where
\begin{equation}
\label{eq:rot_rodriguez}
		[\Omega_{t}] = \begin{bmatrix}
    0 & -\psi_t & \theta_t  \\
    \psi_t & 0 & -\phi_t  \\
    -\theta_t & \phi_t & 0
    \end{bmatrix},
\end{equation}
and $\phi_t$, $\theta_t$, and $\psi_t$ are the $x$-, $y$-, and $z$-axis rotation angles at each time $t$, respectively.
The $z$-axis rotation is treated separately, and thus $\textbf{R}_t$ is modeled without it ($\psi_t = 0$).
For small amount of $x$ and $y$ axis rotation angles, the Eq.~(\ref{eq:rot_exponential}) can be approximated by simply discarding the second and higher orders of the matrix exponential form of Eq.~(\ref{eq:rot_exponential}) ($\textbf{R}_t \approx I + [\Omega_t]$).
This term is written as
\begin{equation}
\label{eq:rot_matrix}
    \textbf{R}_t \approx \begin{bmatrix}
    1 & 0 & \theta_t  \\
    0 & 1 & -\phi_t  \\
    -\theta_t & \phi_t & 1
    \end{bmatrix}.
\end{equation}
The intrinsic matrix $\textbf{K}_u$ is defined by a standard form of
\begin{equation}
\label{eq:intrins_mat}
		\textbf{K}_u = \begin{bmatrix}
		f & 0 & p_u  \\
		0 & f & q_u  \\
		0 & 0 & 1
		\end{bmatrix},
\end{equation}
where $f$ is equal to the focal length and $(p_u, q_u)$ represents the principal point on each sub-aperture image.
Combining Eq.~(\ref{eq:homog_eq}) with matrices of Eq.~(\ref{eq:rot_matrix}) and Eq.~(\ref{eq:intrins_mat}) produces a $3\times3$ homography matrix as follows
\begin{equation}
\label{eq:homog_eq_rotxy}
		\textbf{H}_t = \begin{bmatrix}
		1 & 0 & f\theta_t  \\
		0 & 1 & -f\phi_t  \\
		0 & 0 & 1
		\end{bmatrix}.
\end{equation}
To this end, any spatial points ($x, y$) on the 2D sub-aperture image $S(u)$ of Eq.~(\ref{eq:eq_3DOF_blur}) are translated by the parameters of Eq.~(\ref{eq:homog_eq_rotxy}) without considering their angular location since principal points ($p_u, q_u$) have been eliminated.
The Eq.~(\ref{eq:homog_eq_rotxy}) are remodified by updating the original in-plane translation parameters ($p_x(t), p_y(t)$) as
\begin{align}
\label{eq:rot_to_trans}
    p_x^i(t) &= p_x(t)+f\theta_t, \\
		p_y^i(t) &= p_y(t)-f\phi_t,
\end{align}
where $p_x^i(t)$ and $p_y^i(t)$ are the updated in-plane motion along $x$ and $y$ respectively.
Finally, Eq.~(\ref{eq:eq_3DOF_blur}) is updated using the new parameters from Eq.~(\ref{eq:rot_to_trans}) which can be written as
\begin{equation}
\label{eq:basic_blur_eq_plus_inplane}
		B(x,u) = \int_{t} S(x,u + p_x^i(t) - x p_z(t)) \,dt.
\end{equation}
\begin{figure}
\begin{center}
		\includegraphics[width=0.45\textwidth]{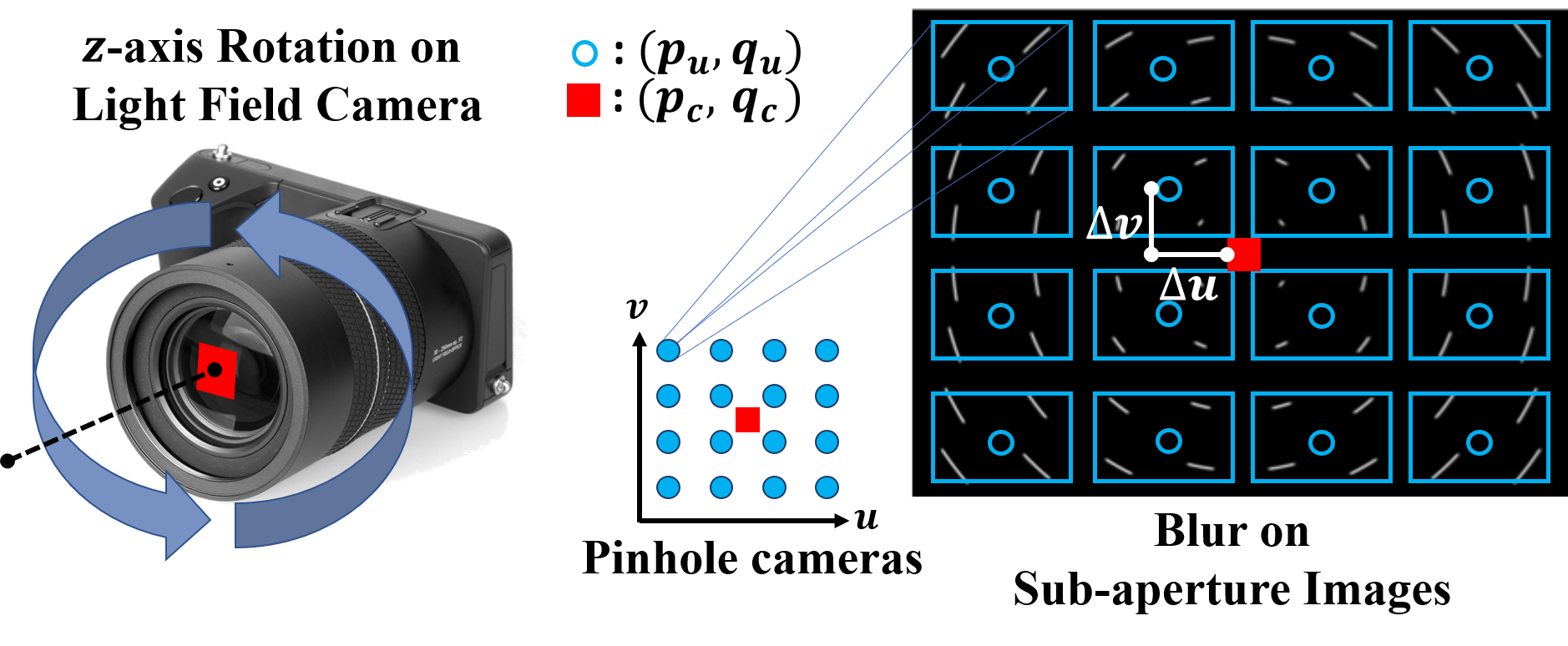}
\end{center}
\vspace{-0.2cm}
\caption{Illustration of the blurred result on sub-aperture images when the LF camera is rotated along the $z$-axis.}
\label{fig:LF_Inplane_Rotation}
\end{figure}
\\
\\
\noindent{\bf In-plane Rotation}
The case of in-plane rotation ($z$-axis rotation), compared with out-of-plane rotation, on the LF camera requires a principal point location.
The reason for this condition is that the $z$-axis intersects at a point in the sub-aperture image plane, which is depicted as a red square in Figure~\ref{fig:LF_Inplane_Rotation}.
For simplicity, we call this red square representation as central angular principal point ($p_c, q_c$).
When this rotation happens, every sub-aperture image is rotated along $p_c, q_c$.
To calculate the exact position of $p_c, q_c$ from each sub-aperture image, we utilize the toolbox of Bok~\etal~\cite{Bok_TPAMI2017}, which provides the LF pinholes' baselines location.
We determine that the sub-aperture image located close to the $p_c, q_c$ is separated with a baseline distance around 0.9 pixel, which was rounded up to 1 pixel for simplicity.
Thus, any sub-aperture image's principal point location ($p_u, q_u$) is separated from the central angular principal point ($p_c, q_c$) with a baseline distance of $\Delta_u, \Delta_v$.
With these parameters having been determined, a simple $z$-axis rotation is performed by plugging the rotation angle ($\psi_t$) into the function that rotates the spatial coordinate ($x, y$).
This function can be interpreted as follows
\begin{align}
\label{eq:update_z_rot}
		x^j &= (x-\Delta_u) \cos (\psi_t)-(y-\Delta_v) \sin (\psi_t) + \Delta_u, \\
		y^j &= (x-\Delta_u) \sin (\psi_t)+(y-\Delta_v) \cos (\psi_t) + \Delta_v.
\end{align}
Combining Eq.~(\ref{eq:update_z_rot}) with Eq.~(\ref{eq:basic_blur_eq_plus_inplane}) produces the full 6-DOF representation, which is written as
\begin{align}
\label{eq:full_6DOF}
		B(x,u) &= \int_{t} S(x^j, u + p_x^i(t) - x^j p_z(t)) \,dt.
\end{align}
Hence, by using the expression in Eq.~(\ref{eq:full_6DOF}), we generate the blurry LF dataset based on the 6-DOF model.
We omit the usage of $y^{j}$ on Eq.~(\ref{eq:full_6DOF}) and $p^{i}_{y}(t)$ on Eq.~(\ref{eq:basic_blur_eq_plus_inplane}) to preserve the parameter consistency from Eq.~(\ref{eq:eq_3DOF_blur}).

\begin{figure*}[t]
\begin{center}
		{\includegraphics[width=0.95\textwidth]{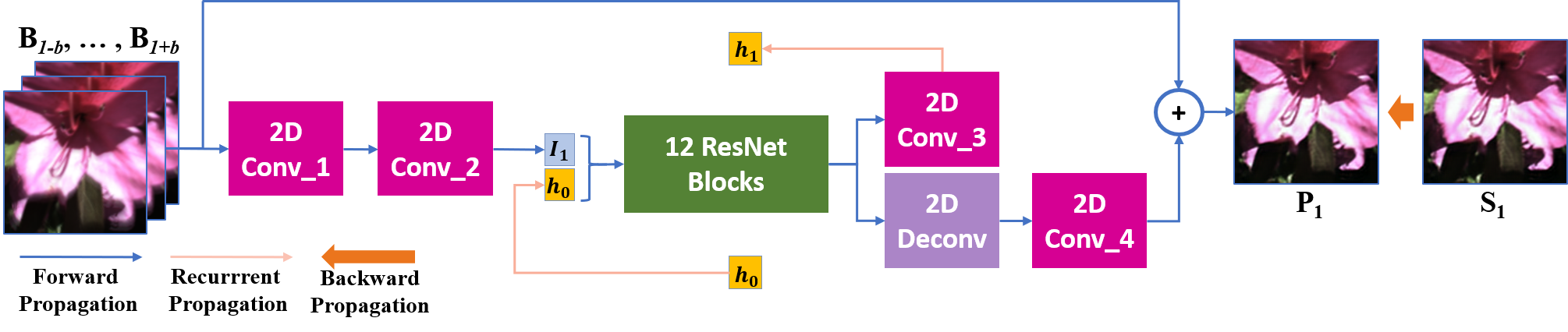}}
\end{center}
\vspace{-0.2cm}
\caption{Our deblurring network with the combined version of simple convolution-deconvolution layers and recurrent procedures.}
\label{fig:Network_Architecture_Fig}
\end{figure*}

\subsection{Light Field Recurrent Deblurring Network}
\label{LFRDBN_sec}
\begin{figure}
\begin{center}
		{\includegraphics[width=0.48\textwidth]{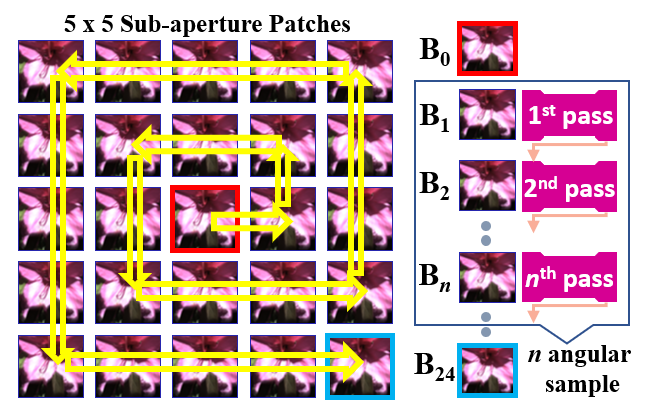}}
\end{center}
\vspace{-0.2cm}
\caption{Spiral stacking and angular sampling are the main strategies of our training. $n$ sub-aperture images are fed into the network multiple times to perform multiple passes.}
\label{fig:Spiral_angular_fig}
\end{figure}

We introduce a learning-based LF deblurring network, denoted as \textit{Light Field Recurrent Deblurring Network} (LFRDBN), which runs in a recurrent manner, as shown in Figure~\ref{fig:Network_Architecture_Fig}.
In general, the angular information of the LF can be regarded as the difference among the sub-aperture images, and we can reconstruct high-quality LF images using such information.
However, as the sub-aperture LF images are degraded by motion blur, accurate angular differences in the blurry images are difficult to calculate.
We tackle this problem by creating a simple network that deblurs each sub-aperture image individually.
The network is designed with the objectives of fast deblurring process, full spatio-angular resolution capability, and high-performance deblurring under complex blur model.
\\
\\
\noindent{\bf Network Architecture}
We use the combination of convolution-deconvolutional and residual styles on the network, which is proven to produce satisfying results on image deblurring~\cite{Kim_ICCV2017, Nah_CVPR2017}.
In particular, our network incorporates the convolution procedure in the first 2 layers, 12 ResNet blocks~\cite{He_CVPR2016}, followed by a parallel function of single deconvolution and convolution layers for recurrent propagation, and provides output via a convolution layer.
Fully convolutional operations are applied to the network, resulting in arbitrary input and output spatial size.
This approach benefits us in achieving a full spatial dimension on LF.
The details of filter, stride, input, output, and channel size are shown in the supplementary material.
We draw on a previous study~\cite{Kim_CVPR2016} by predicting and adding the residual image with the input to produce the deblurred result.
This strategy is supported by applying instance normalization~\cite{Ulyanov_CoRR} and ReLU activation on every 2D convolution layer, except for the last layer \textit{2D\_Conv4}, which utilizes Tanh activation without any normalization.
The Tanh activation rescales feature values between -1.0 and 1.0 and thus, each pixel in the residual image acts as an enhancer or decreaser for the added input image.
The residual blocks follow the traditional ResNet model~\cite{He_CVPR2016} and simple modification is made by applying instance normalization~\cite{Ulyanov_CoRR} instead of batch normalization to normalize the features' contrasts.
\\
\\
\noindent{\bf Recurrent Approach}
\label{Recurrent_sec}
Simultaneously deblurring all sub-aperture images ($u \times v$) with the original spatio-angular size requires significant memory in current graphic card settings.
Instead, we propose a recurrent network that outputs a single sharp sub-aperture image.
To handle large blur, our network takes consecutive multiple sub-aperture images as inputs and propagates the intermediate feature map $\textbf{h}_a$ at angular step $a$ to the input of the network at the next angular position $a+1$ as introduced in the work of Kim~\etal~\cite{Kim_ICCV2017}.
To do so, we initially stack all the sub-aperture images as shown in Figure~\ref{fig:Spiral_angular_fig} and by that, the 4D LF ($x, y, u, v$) is reduced to 3D ($x, y, a$).
Stacking the sub-aperture images in a zig-zag direction produces huge geometry changes between the right-cornered and left-cornered images.
To avoid this problem, we stack them in a spiral direction to minimize the geometric change between horizontal and vertical directions.
Each network uses the input of several consecutive sub-aperture images in each angular position ($\textbf{B}_{a-b}, .., \textbf{B}_{a}, .., \textbf{B}_{a+b}$) and gives output of predicted sub-aperture image $\textbf{P}_a$.
Consequently, features of the consecutive sub-aperture images can be passed through the same network several times and naturally increase the receptive field without computational overhead.
This approach benefits the capability of the network to handle large motion blur while saving on memory, which is ideal for fast LF deblurring.
Moreover, the consistency between consecutive sub-aperture images is enforced by transferring previous hidden feature maps~$\textbf{h}_{a-1}$ on the network.
\\
\\
\noindent{\bf Objective Function}
Our recurrent network is designed in an end-to-end manner.
Thus, backward propagation can be simply done by initially measuring the loss between predicted and ground truth image.
We utilize the L2 loss between the predicted image $\textbf{P}_a$ and ground truth sharp image $\textbf{S}_a$, and also for the weights regularization to avoid overfitting, \textit{i.e.}
\begin{equation}
Loss=\frac{1}{M}\sum_{i}^{M} \| \textbf{S}_{a,i} - \textbf{P}_{a,i}\|^2  +  \lambda \| \textbf{W} \|^2,
\label{equ_loss}
\end{equation}
where $M$ and $\textbf{W}$ denote the number of pixels in each sub-aperture image $a$ ($a \in u \times v$) and  all the trainable parameters, respectively.
$\lambda$  is set to $10^{-4}$ for the regularization function.
\\
\\
\noindent{\bf Angular Sampling}
The angular sampling is simply done by obtaining $n$ of $u \times v$ sub-aperture images under spiral direction as shown in Figure~\ref{fig:Spiral_angular_fig}.
The sampling starts from the center and ends at the right-most bottom part of the angular position.
Angular sampling is beneficial for reducing the training time while also increasing the network's receptive fields by performing multiple passes~\cite{Kim_ICCV2017}.
\section{Experimental Results}
In this section, we quantitatively and qualitatively evaluate the proposed method.
To do so, we initially present the details for acquiring our training dataset, which includes 6-DOF motion blur, to train the proposed network.
We implement a tool for synthesizing LF blur and generate the dataset using MATLAB application.
On the next stage, we build the LF deblurring network by utilizing TensorFlow~\cite{Abadi_OSDI16} library.
Finally, we demonstrate the superiority of the proposed work by comparing with state-of-the-art deblurring methods.
All experiments are performed in a desktop with i7-6700 CPU and NVIDIA GTX 1080 GPU.
\subsection{6-DOF Blur Dataset}
\label{Blur_dataset_sec}
A robust deep learning based algorithm depends on the quality of its training and test datasets.
Recent blur datasets are only available for 2D or 3D image (video) deblurring~\cite{Kim_ICCV2017,Nah_CVPR2017,Su_CVPR2017}.
To the best of our knowledge, no LF dataset includes a non-uniform motion blur caused by 6-DOF camera motion.
Current popular large-scale LF datasets~\cite{Kalantari_TOG2016, Srinivasan_ICCV2017} are provided only for angular super-resolution purposes.
Therefore, we generate a large-scale LF blur dataset by utilizing our real Lytro Illum images and 3D scenes from UnrealCV~\cite{Qiu_ICM2017}.
The synthetic 3D scenes are included in the dataset to tackle the low contrast nature of Lytro Illum LF images.

Our dataset is divided into 360 LF training set and 40 LF test set.
The ratio of real data and 3D scenes are 50:50 in both training and test set.
With these sharp 360 LF training set, we synthesize non-uniformly blurred images by using randomly generated $860$ unique 6-DOF camera motions using the Eq.~(\ref{eq:full_6DOF}) term.
As each LF image includes 25 RGB sub-aperture images $\left(5 \times 5 \times 320 \times 512 \times 3\right)$, we generate $860 \times 25$ pairs of sharp and blurred sub-aperture images for training the network.
Specifically, 430 unique camera trajectories are used to synthetically motion blurred the Illum data while the rest 430 are used in the 3D UnrealCV scenes.
We also set unique camera trajectories on the 40 LF test set for benchmarking the deblurring performance.

To synthesize the motion blur in LF images, we simulate long shutter-time effect by taking the average of differently warped images from the uniformly-sampled versions of 6-DOF camera motion during shutter-time $T$.
Note that, the warped sharp image at time $\frac{T}{2}$ is annotated as the reference sharp LF image in our dataset.
Moreover, to sample the 6-DOF camera motion for warping, we parameterize the 6-DOF motion using Be\'zier curve for translations and spherical interpolation for rotations.
In practice, we trim the original angular resolution size of $13 \times 13$ Illum LF images to $5 \times 5$ to prevent ghosting and vignette results from the 4D LF extraction tool~\cite{Bok_TPAMI2017}.
\begin{figure*}[t]
\begin{center}
		{\includegraphics[width=0.98\textwidth]{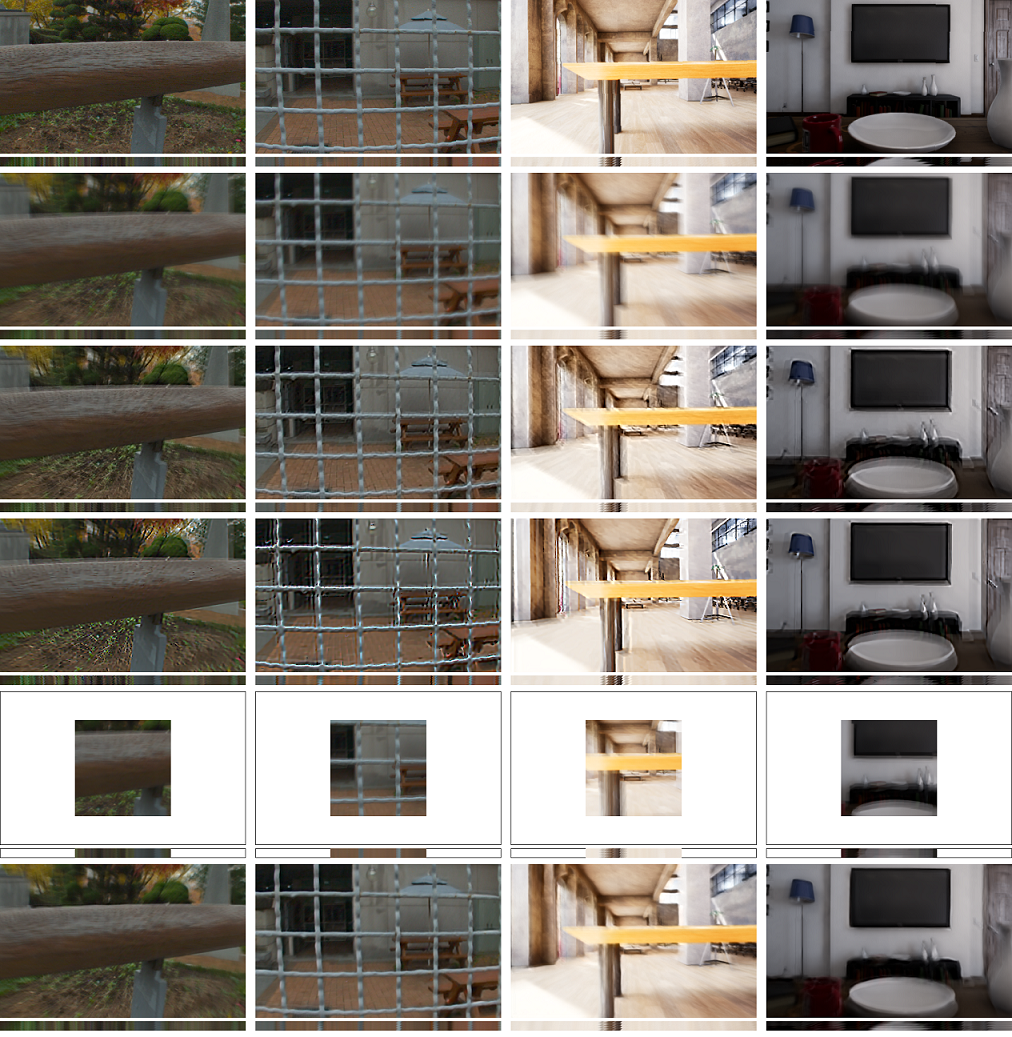}} \\
\end{center}
\vspace{-0.3cm}
\caption{Qualitative comparison from our test set. Images shown above are only the central sub-aperture image of each LF. Each image has the epipolar plane image (EPI) to show the parallax change of a LF. First row images are the original sharp images at time $\frac{T}{2}$. Second row images are the blurry LF that is synthetically made by our 6-DOF model. Third, fourth, fifth, and sixth row images are the deblurred result from the state-of-the-art LF deblurring by Krishnan~\etal~\cite{Krishnan_CVPR2011}, Pan~\etal~\cite{Pan_CVPR2014}, Srinivasan~\etal~\cite{Srinivasan_CVPR2017}, and ours, respectively. The 2 columns on the left are LF images from Illum camera while the 2 on the right are from UnrealCV~\cite{Qiu_ICM2017}.}
\label{fig:Qual_6DOF_Result_fig}
\end{figure*}
\begin{figure*}[t]
\begin{center}
		{\includegraphics[width=0.8\textwidth]{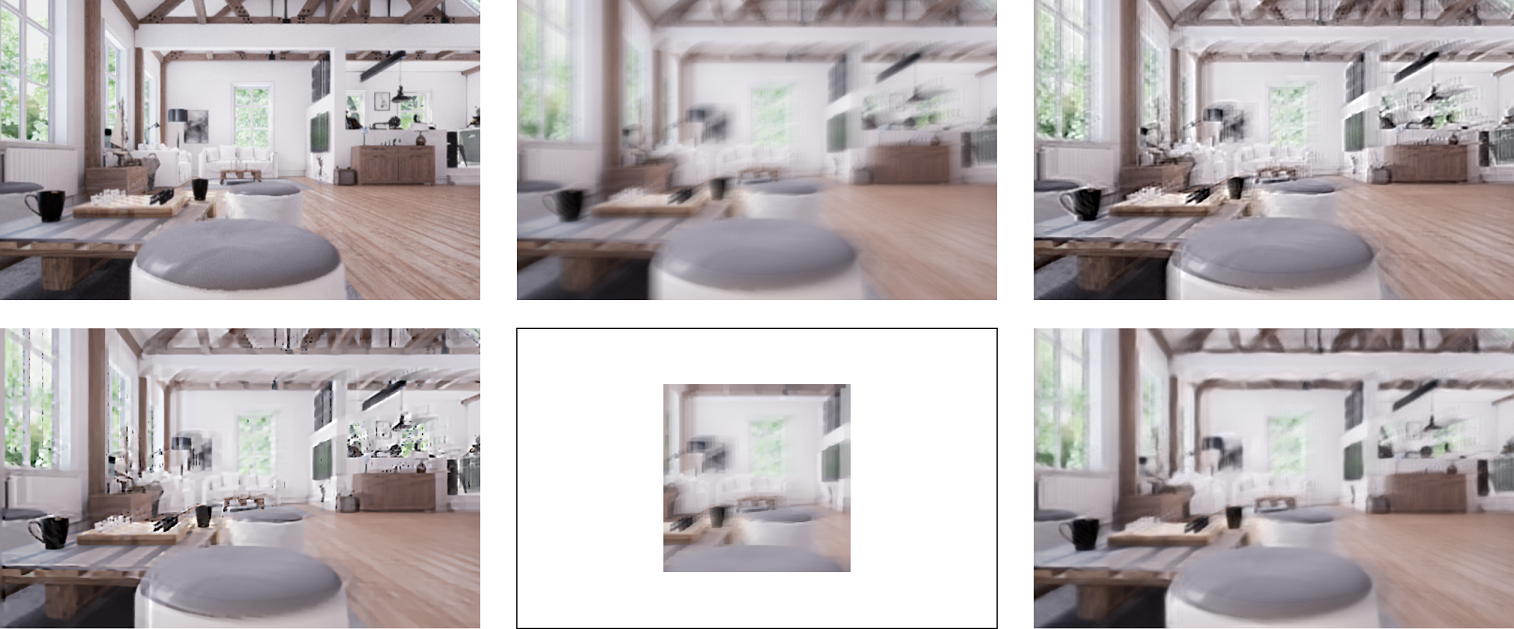}} \\
\end{center}
\vspace{-0.2cm}
\caption{Example of deblurring result using LF data blurred with 3-DOF model~\cite{Srinivasan_CVPR2017}. Top row left to right: sharp input, 3-DOF blurry LF, and deblurred result by Krishnan~\etal~\cite{Krishnan_CVPR2011}, respectively. Bottom row left to right: deblurred result by Pan~\etal~\cite{Pan_CVPR2014}, Srinivasan~\etal~\cite{Srinivasan_CVPR2017}, and ours, respectively. Our network that is trained with 6-DOF model is also capable of deblurring the 3-DOF motion blurred LF.}
\label{fig:Qual_3DOF_Result_fig}
\end{figure*}
\begin{figure*}[t]
\begin{center}
		{\includegraphics[width=0.98\textwidth]{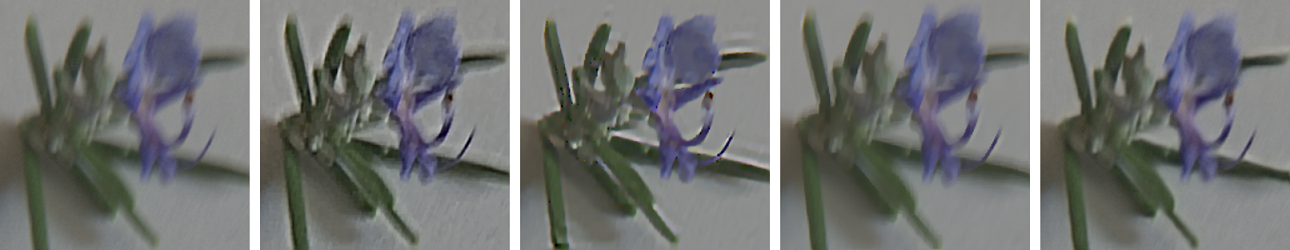}} \\
\end{center}
\vspace{-0.2cm}
\caption{Deblurring results from the real motion blurred LF obtained from the work of Srinivasan~\etal~\cite{Srinivasan_CVPR2017}. From left to right: real blurry LF, deblurring result of Krishnan~\etal~\cite{Krishnan_CVPR2011}, Pan~\etal~\cite{Pan_CVPR2014}, Srinivasan~\etal~\cite{Srinivasan_CVPR2017}, and ours, respectively.}
\label{fig:Qual_Real_Result_fig}
\end{figure*}
\subsection{Network Training}
\label{net_training_sec}
To train the deblurring network, we crop a patch from each sub-aperture image with the size of $256 \times 256$ pixels and they are chosen randomly in the spatial coordinates.
Subsequently, we perform angular sampling with $n = 10$ from the stacked sub-aperture images.
This procedure is equivalent to training our network without building a 10 times deeper version of it.
We set batch = 1 and apply instance normalization for contrast balancing and faster convergence.
Color augmentation is also utilized by switching RGB channels on the cropped patch.
In our implementation, $b$ is set to 1 and thus the network receives 3 consecutive sub-aperture images ($\textbf{B}_{a-1}, \textbf{B}_{a}, \textbf{B}_{a+1}$) for predicting an output sub-aperture image $\textbf{P}_a$.
The network is configured using ADAM optimizer with a constant learning rate of 0.0001 and trained for 400K iterations within 14 hours.
\begin{table}
\begin{center}
\begin{small}
\resizebox{\columnwidth}{!}{
\begin{tabular}{|c|| c | c | c | c|}
\hline
\multirow{ 3}{*}{Method} & \multicolumn{2}{c|}{Cropped LF with 6-DOF Blur } & \multicolumn{2}{c|}{Full LF with 6-DOF Blur}\\
& \multicolumn{2}{c|}{($40\times5\times5\times200\times200\times3$)} & \multicolumn{2}{c|}{($40\times5\times5\times320\times512\times3$) }\\
\cline{2-3}\cline{4-5}
 & PSNR / SSIM & RMSE & PSNR / SSIM & RMSE\\
\hline\hline
~\cite{Krishnan_CVPR2011} &  23.41 / .778 & 0.0732 & 23.41 / .774 & 0.0703\\
\hline
~\cite{Pan_CVPR2014} & 21.70 / .732 & 0.0875 & 21.58 / .730 & 0.0857\\
\hline
~\cite{Srinivasan_CVPR2017} & 23.61 / .765 & 0.0703 &  \multicolumn{2}{c |}{Not available}\\
\hline
Ours & \textbf{26.57} / \textbf{.846} & \textbf{0.0490} & \textbf{25.73} / \textbf{.840} & \textbf{0.0531}\\
\hline
\end{tabular}}
\end{small}
\end{center}
\vspace{-0.3cm}
\caption{PSNR, SSIM, and RMSE result of the deblurring algorithms on processing 6-DOF motion blurred LF from our model.}
\label{tab:Quant_result_6DOF}
\end{table}
\begin{table}
\begin{center}
\begin{small}
\resizebox{\columnwidth}{!}{
\begin{tabular}{|c|| c | c | c | c|}
\hline
\multirow{ 3}{*}{Method} & \multicolumn{2}{c|}{Cropped LF with 3-DOF Blur } & \multicolumn{2}{c|}{Full LF with 3-DOF Blur}\\
& \multicolumn{2}{c|}{($6\times5\times5\times200\times200\times3$)} & \multicolumn{2}{c|}{($6\times5\times5\times320\times512\times3$) }\\
\cline{2-3}\cline{4-5}
 & PSNR / SSIM & RMSE & PSNR / SSIM & RMSE\\
\hline\hline
~\cite{Krishnan_CVPR2011} &  24.50 / .787 & 0.0717 & 25.08 / .809 & 0.0622\\
\hline
~\cite{Pan_CVPR2014} & 21.98 / .731 & 0.0866 & 22.86 / .767 & 0.0769\\
\hline
~\cite{Srinivasan_CVPR2017} & 24.75 / .781 & 0.0673 &  \multicolumn{2}{c |}{Not available}\\
\hline
Ours & \textbf{27.57} / \textbf{.855} & \textbf{0.0453} & \textbf{27.21} / \textbf{.871} & \textbf{0.0459}\\
\hline
\end{tabular}}
\end{small}
\end{center}
\vspace{-0.3cm}
\caption{PSNR, SSIM, and RMSE result of the deblurring algorithms on processing 3-DOF motion blurred LF~\cite{Srinivasan_CVPR2017}.}
\label{tab:Quant_result_3DOF}
\end{table}
\begin{table}
\begin{center}
\begin{small}
\begin{tabular}{|c|| c || c |}
\hline
Method & Cropped LF [sec] & Full LF [sec]\\
\hline\hline
~\cite{Krishnan_CVPR2011} & $\sim$100 (CPU) & $\sim$445 (CPU)\\
\hline
~\cite{Pan_CVPR2014} & $\sim$240 (CPU) & $\sim$1K (CPU)\\
\hline
~\cite{Srinivasan_CVPR2017} & $\sim$8K (GPU) & Not available\\
\hline
Ours & \textbf{$\sim$0.5} (GPU) & \textbf{$\sim$1.7} (GPU)\\
\hline
\end{tabular}
\end{small}
\end{center}
\vspace{-0.3cm}
\caption{Execution time comparison between state-of-the-art algorithms and our method. Each cell represents the algorithm's runtime of deblurring total $5 \times 5$ sub-aperture images in a LF.}
\label{tab:Time_result_tab}
\end{table}
\subsection{Quantitative Results}
We compare our result with the state-of-the-art LF deblurring algorithm by Srinivasan~\etal~\cite{Srinivasan_CVPR2017} that is publicly available.
For more objective evaluations, we also perform the comparison with 2D image deblurring algorithms by Krishnan~\etal~\cite{Krishnan_CVPR2011} and Pan~\etal~\cite{Pan_CVPR2014} as recommended by Lai~\etal~\cite{Lai_CVPR2016}.
The 2D deblurring algorithms are applied individually on each sub-aperture image.
We utilize the peak signal-to-noise ratio (PSNR), structural similarity index (SSIM), and root mean-squared error (RMSE) functions to compare the performance in numbers.
In Table~\ref{tab:Quant_result_6DOF} and~\ref{tab:Quant_result_3DOF}, each cell's value is achieved by averaging all PSNRs, SSIMs, and RMSEs from the 40 LF test set from our 6-DOF model and smaller set of 6 LF for 3-DOF model.
We provide exactly 10~\% of the dataset for testing procedure in the 6-DOF comparisons while using a smaller set of 3-DOF motion blurred data to show our advantage of the 6-DOF model over the 3-DOF model.
Note that bigger PSNR and SSIM numbers represent better result, while smaller RMSE number represents a better result.
As shown in Table~\ref{tab:Quant_result_6DOF}, our network produces the best result among the competitors in terms of the 6-DOF blurry test set LF.
Another aspect of our method is that although it is trained on the full 6-DOF model, it achieves state of the art performance on the special case of purely translational 3-DOF motion blurred data.  This comparison is shown in Table~\ref{tab:Quant_result_3DOF}.
Note that this is a task for which Srinivasan~\etal~\cite{Srinivasan_CVPR2017} is specifically designed.

Table~\ref{tab:Time_result_tab} shows the superiority of our network in terms of execution speed where full spatio-angular LF deblurring is performed in less than 2 seconds and its cropped version in less than 1 second.
Each value in Table~\ref{tab:Time_result_tab} represents the total duration of each algorithm for processing $5 \times 5$ sub-aperture images.
Our method surpasses others' by the speed of $\sim$15 frames per second rate ($320 \times 512$)  using the recent NVIDIA GTX 1080 graphic card.
Specifically, our method able to process 4D LF 16,000 times faster than the state-of-the-art LF deblurring~\cite{Srinivasan_CVPR2017} with the same GPU setting.
\subsection{Qualitative Results}
We provide visual results for measuring the deblurring performance qualitatively.
Deblurring performance on our test set is shown in Figure~\ref{fig:Qual_6DOF_Result_fig} where the 2 columns on the left are the examples of Lytro Illum LF data and the 2 columns on the right are the examples of UnrealCV 3D scenes~\cite{Qiu_ICM2017}.
More details about the blurry, sharp ground truth, and deblurring results of Figure~\ref{fig:Qual_6DOF_Result_fig} are provided on the captions.

Visually, our results show better performance than others' especially on the large region with large motion blur.
As for the 3-DOF comparison, visual result is shown in the Figure~\ref{fig:Qual_3DOF_Result_fig}.
To show the robustness of our network, deblurring is performed on real blurry LF data obtained from Srinivasan~\etal~\cite{Srinivasan_CVPR2017} and a visual example of the performance is shown in Figure~\ref{fig:Qual_Real_Result_fig}.
Furthermore, to show the generality of our network, we include the result on the 6-DOF blurred LF data that is collected from another dataset~\cite{Kalantari_TOG2016} which is totally different with our training and test set.
This result can be seen in Figure~\ref{fig:Show_up_fig} in the beginning.
Epipolar plane images (EPI) are also included to show the LF consistency.
We encourage the readers to see them on electronic screen or simply watch our video in the supplementary material.
\section{Conclusion}
In this work, we presented a LF deblurring method that is performed via neural network.
The network is employed with recurrent approach to achieve large receptive field capability without building a very deep architecture.
To train the network, we synthesized blurry LF dataset based on the 6-DOF motion model on LF camera.
Our network was crafted to process full 4D LF input and output and was trained in an effective manner.
Our method outperformed the state-of-the-art LF deblurring in terms of deblurring result, full resolution, and fast processing capability.
{\small
\bibliographystyle{ieee}
\bibliography{egbib}

\begin{thebibliography}{10}\itemsep=-1pt

\bibitem{Lytro1}
Lytro. {T}he {L}ytro {C}amera.

\bibitem{Raytrix1}
Raytrix - 3D {L}ight {F}ield {C}amera {T}echnology.

\bibitem{Abadi_OSDI16}
M.~Abadi, P.~Barham, J.~Chen, Z.~Chen, A.~Davis, J.~Dean, M.~Devin,
  S.~Ghemawat, G.~Irving, M.~Isard, et~al.
\newblock Tensorflow: a system for large-scale machine learning.
\newblock In {\em Proc. of the 12th USENIX Conference on Operating Systems
  Design and Implementation}, volume~16, pages 265--283, 2016.

\bibitem{Bok_TPAMI2017}
Y.~Bok, H.-.~G. Jeon, and I.~S. Kweon.
\newblock Geometric calibration of micro-lense-based light field cameras using
  line features.
\newblock {\em IEEE Trans. on Pattern Analysis and Machine Intelligence},
  39(2):287--300, 2017.

\bibitem{Dansereau_CVPRW2017}
D.~G. Dansereau, A.~Eriksson, and J.~Leitner.
\newblock Richardson-{L}ucy deblurring for moving light field cameras.
\newblock In {\em Proc. of IEEE Conference on Computer Vision and Pattern
  Recognition Workshop}, pages 70--81, 2017.

\bibitem{Gupta_ECCV2010}
A.~Gupta, N.~Joshi, C.~L. Zitnick, M.~Cohen, and B.~Curless.
\newblock Single image deblurring using motion density functions.
\newblock In {\em Proc. of European Conference on Computer Vision}, pages
  171--184. Springer, 2010.

\bibitem{He_CVPR2016}
K.~He, X.~Zhang, S.~Ren, and J.~Sun.
\newblock Deep residual learning ofr image recognition.
\newblock In {\em Proc. of IEEE Conference on Computer Vision and Pattern
  Recognition}, pages 770--778, 2016.

\bibitem{Kalantari_TOG2016}
N.~K. Kalantari, T.-C. Wang, and R.~Ramamoorthi.
\newblock Learning-based view synthesis for light field cameras.
\newblock {\em ACM Transactions on Graphics}, 35(6), 2016.

\bibitem{Kim_CVPR2016}
J.~Kim, J.~K. Lee, and K.~M. Lee.
\newblock Accurate image super-resolution using very deep convolutional
  networks.
\newblock In {\em Proc. of IEEE Conference on Computer Vision and Pattern
  Recognition}, pages 1646--1654, 2016.

\bibitem{Kim_ICCV2017}
T.~H. Kim, K.~M. Lee, B.~Scholkopf, and M.~Hirsch.
\newblock Online video deblurring via dynamic temporal blending network.
\newblock In {\em Proc. of IEEE International Conference on Computer Vision},
  pages 4058--4067, 2017.

\bibitem{Krishnan_CVPR2011}
D.~Krishnan, T.~Tay, and R.~Fergus.
\newblock Blind deconvolution using a normalized sparsity measure.
\newblock In {\em Proc. of IEEE Conference on Computer Vision and Pattern
  Recognition}, pages 233–--240, 2011.

\bibitem{Kupyn_CVPR2018}
O.~Kupyn, V.~Budzan, M.~Mykhailych, D.~Mishkin, and J.~Matas.
\newblock Deblurgan: Blind motion deblurring using conditional adversarial
  networks.
\newblock In {\em Proc. of IEEE Conference on Computer Vision and Pattern
  Recognition}, pages 8183--8192, 2018.

\bibitem{Lai_CVPR2016}
W.~S. Lai, J.~B. Huang, Z.~Hu, N.~Ahuja, and M.~H. Yang.
\newblock A comparative study for single image blind deblurring.
\newblock In {\em Proc. of IEEE Conference on Computer Vision and Pattern
  Recognition}, pages 1701--1709, 2016.

\bibitem{Dongwoo_ECCV2018}
D.~Lee, H.~Park, I.~K. Park, and K.~M. Lee.
\newblock Joint blind motion deblurring and depth estimation of light field.
\newblock In {\em Proc. of European Conference on Computer Vision}, 2018.

\bibitem{Levin_TOG2007}
A.~Levin, R.~Fergus, F.~Durand, and W.~T. Freeman.
\newblock Image and depth from a conventional camera with a coded aperture.
\newblock {\em ACM Trans. on Graphics}, 26(3):70, 2007.

\bibitem{Mohan_CVPR2018}
M.~Mahesh~Mohan and A.~Rajagopalan.
\newblock Divide and conquer for full-resolution light field deblurring.
\newblock In {\em Proc. of the IEEE Conference on Computer Vision and Pattern
  Recognition}, pages 6421--6429, 2018.

\bibitem{Nah_CVPR2017}
S.~Nah, T.~H. Kim, and K.~M. Lee.
\newblock Deep multi-scale convolutional neural network for dynamic scene
  deblurring.
\newblock In {\em Proc. of the IEEE Conference on Computer Vision and Pattern
  Recognition}, volume~1, page~3, 2017.

\bibitem{Ng_TOG2015}
R.~Ng.
\newblock Fourier slice photography.
\newblock {\em ACM Trans. on Graphics}, 24(3):735--744, 2005.

\bibitem{Pan_CVPR2014}
J.~Pan, Z.~Hu, Z.~Su, and M.~H. Yang.
\newblock Deblurring text images via ${L}_0$-regularized intensity and gradient
  prior.
\newblock In {\em Proc. of IEEE Conference on Computer Vision and Pattern
  Recognition}, pages 2901--2908, 2014.

\bibitem{Pan_CVPR2016}
J.~Pan, D.~Sun, H.~Pfister, and M.~H. Yang.
\newblock Blind image deblurring using dark channel prior.
\newblock In {\em Proc. of IEEE Conference on Computer Vision and Pattern
  Recognition}, pages 1628--1636, 2016.

\bibitem{Qiu_ICM2017}
W.~Qiu, F.~Zhong, Y.~Zhang, S.~Qiao, Z.~Xiao, T.~S. Kim, and Y.~Wang.
\newblock Unrealcv: Virtual worlds for computer vision.
\newblock In {\em Proc. of the 25th ACM International Conference on
  Multimedia}, pages 1221--1224, 2017.

\bibitem{Srinivasan_CVPR2017}
P.~P. Srinivasan, R.~Ng, and R.~Ramamoorthi.
\newblock Light field blind motion deblurring.
\newblock In {\em Proc. of IEEE Conference on Computer Vision and Pattern
  Recognition}, pages 3958--3966, 2017.

\bibitem{Srinivasan_ICCV2017}
P.~P. Srinivasan, T.~Wang, A.~Sreelal, R.~Ramamoorthi, and R.~Ng.
\newblock Learning to synthesize a 4d rgbd light field from a single image.
\newblock In {\em IEEE International Conference on Computer Vision}, pages
  2262--2270, 2017.

\bibitem{Su_CVPR2017}
S.~Su, M.~Delbracio, J.~Wang, G.~Sapiro, W.~Heidrich, and O.~Wang.
\newblock Deep video deblurring for hand-held cameras.
\newblock In {\em Proc. of IEEE Conference on Computer Vision and Pattern
  Recognition}, volume~2, page~6, 2017.

\bibitem{Tai_TPAMI2011}
Y.~W. Tai, P.~Tan, and M.~S. Brown.
\newblock Richardson-lucy deblurring for scenes under a projective motion path.
\newblock {\em IEEE Trans. on Pattern Analysis and Machine Intelligence},
  33(8):1603--1618, 2011.

\bibitem{Ulyanov_CoRR}
D.~Ulyanov, A.~Veldadi, and V.~S. Lempitsky.
\newblock Instance normalization: the missing ingredient for fast stylization.
\newblock In {\em CoRR abs/1607.08022}, 2016.

\bibitem{Wang_TPAMI2016}
T.-C. Wang, A.~Efros, and R.~Ramamoorthi.
\newblock Depth estimation with occlusion modeling using light-field cameras.
\newblock {\em IEEE Trans. on Pattern Analysis and Machine Intelligence},
  38(11):2170--2181, 2016.

\bibitem{Whyte_CVPR2010}
O.~Whyte, J.~Sivic, A.~Zisserman, and J.~Ponce.
\newblock Non-uniform deblurring for shaken images.
\newblock In {\em Proc. of IEEE Conference on Computer Vision and Pattern
  Recognition}, number~2, pages 168--186, 2010.

\bibitem{Williem_TPAMI2018}
W.~Williem, I.~K. Park, and K.~M. Lee.
\newblock Robust light field depth estimation using occlusion-noise aware data
  costs.
\newblock {\em IEEE Trans. on Pattern Analysis and Machine Intelligence},
  40(10):2484--2497, 2018.

\bibitem{Xu_ECCV2010}
L.~Xu and J.~Jia.
\newblock Two-phase kernel estimation for robust motion deblurring.
\newblock In {\em Proc. of European Conference on Computer Vision}, pages
  157--170, 2010.

\end{thebibliography}
}

\end{document}